\documentclass[11pt,a4paper]{article}
\PassOptionsToPackage{hyphens}{url}\usepackage{hyperref}
\usepackage[hyperref]{AACL-IJCNLP2020}
\usepackage{times}
\usepackage{latexsym}

\usepackage{microtype}

\usepackage{comment}
\aclfinalcopy 

\usepackage[utf8]{inputenc}
\usepackage[T4,T1]{fontenc}

\newcommand{\bam}[1]{{\fontencoding{T4}\selectfont#1}}

\usepackage{multirow}
\usepackage{textcomp}
\usepackage{graphicx} 
\usepackage{lscape}
\usepackage{booktabs}
\usepackage{wrapfig}
\usepackage{amsmath} 
\usepackage{amsfonts} 
\usepackage{multicol} 

\usepackage{microtype}

\title{Neural Machine Translation for Extremely Low-Resource African Languages: A Case Study on Bambara}

\author{\textbf{Allahsera Auguste Tapo$^{1,\ast}$,
Bakary Coulibaly$^{2}$, Sébastien Diarra$^{2}$, Christopher Homan$^{1}$,}\\ \vspace{2mm} \textbf{Julia Kreutzer$^{3}$, Sarah Luger$^{4}$, Arthur Nagashima$^{1}$, Marcos Zampieri$^{1}$, Michael Leventhal$^{2}$}\\

${^1}$Rochester Institute of Technology\\ 
${^2}$Centre National Collaboratif de l'Education en Robotique et en Intelligence Artificielle (RobotsMali)\\ \vspace{2mm}
${^3}$Google Research, ${^4}$Orange Silicon Valley \\
\texttt{$^\ast$aat3261@rit.edu} }

\date{}

\begin{document}
\maketitle
\begin{abstract}
Low-resource languages present unique challenges to (neural) machine translation. We discuss the case of Bambara, a Mande language for which training data is scarce and requires significant amounts of pre-processing. More than the linguistic situation of Bambara itself, the socio-cultural context within which Bambara speakers live poses challenges for automated processing of this language. In this paper, we present the first parallel data set for machine translation of Bambara into and from English and French and the first benchmark results on machine translation to and from Bambara. We discuss challenges in working with low-resource languages and propose strategies to cope with data scarcity in low-resource machine translation (MT).  
\end{abstract}

\section{Introduction}
Underresourced languages, from a natural language processing (NLP) perspective, are those lacking the resources (large volumes of parallel bitexts) needed to support state-of-the-art performance on NLP problems like machine translation, automated speech recognition, or named entity recognition. Yet the vast majority of the world's languages---representing billions of native speakers worldwide---are underresourced. And the lack of available training data in such languages usually reflects a broader paucity of electronic information resources accessible to their speakers.


For instance, there are over six million Wikipedia articles in English but fewer than sixty thousand in Swahili and fewer than seven hundred in Bambara, the vehicular and most widely-spoken native language of Mali that is the subject of this paper.\footnote{\url{meta.wikimedia.org/wiki/List_of_Wikipedias.}} Consequently, only 53\% of the worlds population have access to ``encyclopedic knowledge'' in their primary language, according to a 2014 study by Facebook.\footnote{\url{fbnewsroomus.files.wordpress.com/2015/02/state-of-connectivity1.pdf}} MT technologies could help bridge this gap, and there is enormous interest in such applications, ironically enough, from speakers of the languages on which MT has thus far had the least success. There is also great potential for humanitarian response applications \cite{ktem2020tigrinya}.

Fueled by data, advances in hardware technology, and deep neural models, machine translation (NMT) has advanced rapidly over the last ten years.  
Researchers are beginning to investigate the effectiveness of (NMT)  low-resource languages, as in recent WMT 2019 and WMT 2020 tasks \cite{barrault2019findings}, and in underresourced African languages.
Most prominently, the \emph{Masakhane} \citep{masakhane} community\footnote{\url{masakhane.io}}, a grassroots initiative, has developed open-source NMT models for over 30 African languages on the base of the JW300 corpus~\citep{agic-vulic-2019-jw300}, a parallel corpus of religious texts. 

Since African languages cover a wide spectrum of linguistic phenomena and language families \cite{heine2000african}, individual development of translations and resources for selected languages or language families are vital to drive the overall progress. Just within the last year, a number of dedicated studies have significantly improved the state of African NMT: \citet{biljon2020} analyzed the depth of Transformers specifically for low-resource translation of South-African languages, based on prior studies by 
\citet{DBLP:journals/corr/abs-1906-05685} on the Autshumato corpus~\citep{autshumato}. \citet{dossou2020ffr} developed an MT model and compiled resources for translations between Fon and French, \citet{akinfaderin-2020-hausamt} modeled translations between English and Hausa, \citet{orife2020neural} for four languages of the Edoid language family, and \citet{ahia2020supervised} investigated supervised vs. unsupervised NMT for Nigerian Pidgin. 


In this paper, we present the first parallel data set for machine translation of Bambara into and from English and French and the first benchmark results on machine translation to and from Bambara. We discuss challenges in working with low-resource languages and propose strategies to cope with data scarcity in low-resource MT.  
We discuss the socio-cultural context of Bambara translation and its implications for model and data development. Finally, we analyze our best-performing neural models with a small-scale human evaluation study and give recommendations for future development.
We find that the translation quality on our in-domain data set is acceptable, which gives hope for other languages that have previously fallen under the radar of MT development.
    
We released our models and data upon publication\footnote{\url{https://github.com/israaar/mt_bambara_data_models}}. Our evaluation setup may serve as benchmark for an extremely challenging translation task.

\section{The Bambara Language}
Bambara is the first language of five million people and the second language of approximately ten million more. Most of its speakers are members of Bambara ethnic groups, who live  throughout the African continent. Approximately 30--40 million people speak some language in the Mande family of languages, to which Bambara belongs~\cite{lewis2014ethnologue}.

Bambara is a tonal language with a rich morphology. Over the years, several competing writing systems have developed, however, as an historically predominately oral language, a majority of Bambara speakers have never been taught to read or write the standard form of the language. Many are incapable of reading or writing the language at all.
The standardization of words and the coinage of new ones are still works in progress; this poses challenges to automated text processing.

During Muslim expansion and French colonization, Arabic and French mixed with local languages, resulting in a lingua franca, e.g., Urban Bambara. Most of the existing Bambara resources are cultural (folk stories or news/topical) or come from social media or text messages, and these are a written in a melange of French, Bambara and Arabic. Consequently, corpora based on common Bambara usage must account for the code switching found in these mixtures.

  Most of these characteristics are shared with related languages, e.g., a subset of the Mande family of languages, where many languages are mutually intelligible. Thus, our hope is that our approach will be transferable to the other twelve official local languages of Mali, or to other African languages with a comparable socio-cultural and linguistic embedding, for example Wolof (non-Mande), which is comparable in terms of number of speakers, borrowings from Arabic and French influence, and oral traditions.
  
 The next section will provide more details on digital resources and describe the process of exploring and collecting data and choosing parallel corpora for the training of the NMT model.

\section{Data Collection}
\subsection{Bambara Corpora}

We discovered that there has been no prior development of automatic translation of Bambara, despite a relatively large volume of research on the language~\citep{culy1985complexity,aplonova2017towards, aplonova2018development}. As a pilot study for assessing the potential for automatic translation of Bambara, \citet{leventhal2020assessing} crowdsourced a small set of written or oral translations from French to Bambara. Additional work was carried out exploring novel crowdsourcing strategies for data collection in Mali \cite{luger2020}. 

The \emph{Corpus Bambara de Référence}~\cite{vydrin2011corpus} is the largest collection of electronic texts in Bambara. It includes scanned and text-based electronic formats. A number of parallel texts based on this data exist. For example, \citet{vydrin2018corpus} analyzed Bambara's separable adjectives using this data.

To survey the known available sources of parallel texts with Bambara, we consulted with a number of authorities on Bambara, including the \emph{Academie Malienne des Langues} (AMALAN) in Mali and the \emph{Institut National des Langues et Civilisations Orientales} (INALCO) in France, as well as a number of individual linguists and machine translation experts throughout the world. These two organisations play key roles in the definition and the promotion of a standard form of written Bambara through the collection and annotation of corpora, the publishing of dictionaries, and, in formulating recommendations for language policy in Mali.

Our efforts uncovered several sources of parallel texts between Bambara and French and/or English that are listed in Table \ref{tab:data_sources} in Appendix~\ref{sec:appendix}. The table provides a rating of each of the identified resources, and the rationale why they were in- or excluded from our translation study. 
  Ultimately, most of these resources proved either of very little or no practical use as sources of training data.
  Many did not actually contain aligned texts and some not even suitable monolingual text. 

    A systematic problem was lack of adherence to the standardized Bambara orthography, due to it being a predominately oral language. This is also one of the reasons why our search for parallel data on the web generally did not yield many finds--- commonly being used in written form, Bambara is used even less on the web. For example, the Bambara Wikipedia contains currently 667 articles (compared to 6M for English), of which a large percentage are only stubs. Of the small number of full articles, most do not consistently employ the standard orthography of Bambara. A selection of those, however, was prepared to be used as monolingual data for MT data augmentation.

Most African NMT studies have been based on the \emph{JW300} corpus \cite{agic-vulic-2019-jw300}, e.g. most of the Masakhane benchmarks ~\citep{masakhane}. \emph{JW300} only contains less than 200 sentences of Dyula, a closely related language to Bambara that it is mutually intelligible with. It might be useful for future cross-lingual studies \citep{wu2019language,goyal-etal-2020-efficient}, but in order to avoid interference between languages, we focus on Bambara data exclusively in this first study.



\begin{figure*}[!h]
    \centering
    \includegraphics[width=\textwidth]{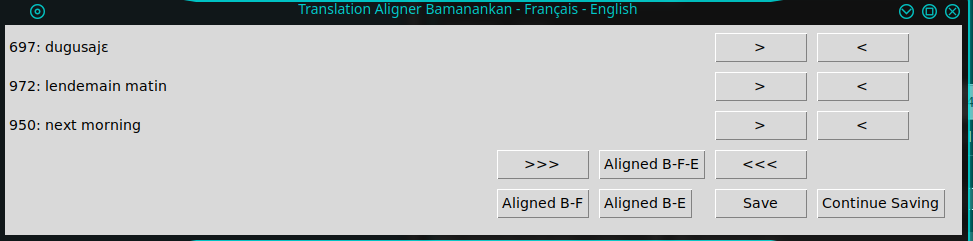}
    \caption[Custom aligner.]{The custom aligner we developed to manually align the dictionary data set. The controls are as follows: for each language, ``$>$'' goes to the next item and ``$<$'' goes to the previous item; for all languages, ``$>>>$'' goes to the next items and ``$<<<$'' goes to the previous items; ``Aligned B-F-E'' saves to memory the alignment of all 3 languages; ``Aligned B-F'' saves to memory the alignment of Bambara and French items; ``Aligned B-E'' saves to memory the alignment of Bambara and English items; ``Save'' saves to a new file; ``Continue Saving'' continues saving the file created.}
    \label{fig:custom_aligner}
\end{figure*}

The most promising for our NMT approach was a dictionary data set from \emph{SIL Mali}\footnote{\url{https://www.sil-mali.org/en/content/introducing-sil-mali}} with examples of sentences used to demonstrate word usage in Spanish, French, English, and Bambara; and a tri-lingual health guide titled ``Where there is no doctor.\footnote{\url{https://gafe.dokotoro.org/}}'' Detailed corpus statistics are listed in Table~\ref{tab:data_table}.

\begin{table}[!h]
\begin{tabular}{ccccc}
\toprule
& & \textbf{Bambara} & \textbf{French} & \textbf{English} \\ \midrule
 \multirow{3}{*}{\rotatebox[origin=l]{90}{\textbf{Dict.}}} & glosses & 3,548 & 4,847 & 4,855 \\ 
 & examples & 2,023 & 2,021 & 2,021 \\ 
 & \textbf{aligned} & 2,158 & 2,146 & 2,158 \\ \midrule
\multirow{7}{*}{\rotatebox[origin=l]{90}{\textbf{Medical}}}  & chapters & \multicolumn{3}{c}{27} \\ 
 & files & \multicolumn{3}{c}{336} \\ 
 & paragraphs & 9,336 & 9,367 & 9,356 \\ 
  & unigrams & 8,209 & 9,893 & 6,935 \\ 
  & bigrams & 26,430 & 25,746 & 31,412 \\ 
 & trigrams & 5,816 & 11,312 & 21,398 \\ 
  & stopwords & 147 & 123 & 69 \\ 
 \bottomrule
\end{tabular}
\caption[Main data-sets.]{The dictionary data set from \emph{SIL Mali} and the medical health guide ``Where there is no doctor'' with examples in French, English, and Bambara. }
\label{tab:data_table}
\end{table}

\subsection{Sentence Alignment}\label{preprocessing}
The part of the dictionary that we are focusing on in this study, are the dictionary entries that consist of examples of Bambara expressions followed by their translations in French and in English. Most of these are single sentences, so there is sentence-to-sentence alignment in the majority of cases. However, there remains a sufficient number of exceptions to render automated pairing impossible. Part of the problem lies in the unique linguistic and cultural elements of the bambaraphone environment; it is often not possible to meaningfully translate an expression in Bambara without giving an explanation of the context.

The medical health guide is aligned by chapters, each of which is roughly aligned by paragraphs. But at the paragraph level there are too many exceptions for automated pairing to be feasible. Many of the bambaraphone-specific problems found in the dictionary dataset are present at the sentence level as well, particularly in explanations of concepts that can be succinctly expressed in English or French but for which Bambara lacks terminology and the bambaraphone environment lacks an equivalent physical or cultural context. 

Both datasets therefore required manual alignment by individuals fluent in written Bambara and either French or English. The annotators need to be able to exercise expert-level judgment on linguistic and, occasionally, medical questions. Access to such human resources was a major factor limiting the quantity of data we were able to align. Because of this, and since the dictionary data was more closely aligned at the sentence level and did not require as much domain knowledge as the medical dataset, we have thus far only used the dictionary dataset in our machine learning experiments.

 In order to facilitate this alignment, we implemented an alignment interface, as shown in Figure \ref{fig:custom_aligner}. 
 It allows annotators to manually align sentences and to save those sentence pairs that another annotator considered properly aligned. In separate tasks, four annotators with a secondary school level understanding of Bambara performed alignment on French-Bambara and English-Bambara sentence pairs using the tool. 


\subsection{Preprocessing}
Before we could align these sentences, we needed to clean the retrieved dictionary entries. 
Below we give examples of cases we had to handle manually, going through the entire corpus line by line.

\begin{enumerate}
    \item Only one language is represented: Discarded.
    \item Ambiguous pronouns in Bambara:\footnote{One can imagine that translating these into French or English is difficult since there is no indicator of the correct choice. \citep{johnson_2018}}\\
    \emph{Before:}\\
    fr: ``Il/elle est né à Bamako en 1938.''\\	
    bam: ``\bam{A bangera Bamak\m{o} san 1938.}'' \\
    \emph{After:}\\
   fr: ``Il est né à Bamako en 1938.''\\ 
   bam: ``\bam{A bangera Bamak\m{o} san 1938.}'' \\
   \emph{and}\\
    fr: ``Elle est né à Bamako en 1938.''\\	
    bam: ``\bam{A bangera Bamak\m{o} san 1938.}''
    \item Additional explanations in the other languages while those are absent in Bambara: \\
    \emph{Before:} \\
    fr: ``Un doigt ne peut pas prendre un caillou (C'est important d'aider les uns les autres).''\\	bam: ``\bam{Bolok\m{o}ni kelen t\m{e} se ka b\m{e}l\m{e} ta.}'' \\
    \emph{After:} \\
    fr: ``Un doigt ne peut pas prendre un caillou.''\\	
    bam: ``\bam{Bolok\m{o}ni kelen t\m{e} se ka b\m{e}l\m{e} ta.} ''
    \item Proverbs: \\
    \emph{Before:}\\
    fr: ``Proverbe: Une longue absence vaut mieux qu'un communiqué (d'un décès).'' \\
    bam: ``\bam{Fama ka fisa ni k\m{o}munike ye.}'' \\
   \emph{After:}\\
    fr: ``Une longue absence vaut mieux qu'un communiqué.''\\	
    bam: ``\bam{Fama ka fisa ni k\m{o}munike ye.}''
\end{enumerate}





Data preparation, including alignment, proved to be about 60\% of the overall time spent in person-hours on the experiment and required on-the-ground organisation and recruitment of skilled volunteers in Mali.
 
\subsection{Parallel Data}
The final data set contains 2,146  parallel sentences of Bambara-French and 2,158  parallel sentences of Bambara-English--a very small data set for NMT compared to the massive state-of-the-art models that are trained on billions of sentences \citep{arivazhagan2019massively}.
We split the data randomly into training, validation, and test sets of 75\%, 12.5\% and 12.5\% respectively. 
The training set is composed of 1611 sentences, the validation set of 268 sentences, the test set of 267 sentences for Bambara-French. The training set is composed of 1620 sentences, the validation set of 270 sentences, the test set of 268 sentences for Bambara-French. 

\subsection{Monolingual Data}
In addition to the translations, we obtained a dataset of 488 monolingual Bambara sentences, sampled from all articles in the Bambara Wikipedia and covering a range of topics, but with preponderance of articles related to Mali. We used  this monolingual dataset for experiments in data augmentation through back-translation, described in Section \ref{sec:auto-eval}.

\section{NMT Development}

\subsection{Hyperparameters}
Our NMT is a transformer~\citep{vaswani2017attention} of appropriate size for a relatively smaller training dataset~\citep{biljon2020}. It has six layers with four attention heads for encoder and decoder, the transformer layer has a size of 1024, and the hidden layer size 256, the embeddings have 256 units. Embeddings and vocabularies are not shared across languages, but the softmax layer weights are tied to the output embedding weights. The model is implemented with the Joey NMT framework~\citep{kreutzer-etal-2019-joey} based on PyTorch \citep{NIPS2019_9015}.

Training runs for 120 epochs in batches of 1024 tokens each. The ADAM optimizer \citep{kingma2014adam} is used with a constant learning rate of 0.0004 to update model weights. This setting was found to be best to tune for highest BLEU, compared to decaying or warmup-cooldown learning rate scheduling.
For regularization, we experimented with dropout and label smoothing~\citep{szegedy2016rethinking}. The best values were 0.1 for dropout and 0.2 for label smoothing across the board.
For inference, beam search with width of 5 is used. The remaining hyperparameters are documented in the Joey NMT configuration files that we will provide with the code.

\subsection{Segmentation}

There is no standard tokenizer for Bambara. Therefore, we simply apply whitespace tokenization for word-based NMT models and compute BLEU with "international" tokenization.\footnote{Tokenizing on punctuation and special symbols, except when surrounding a digit.}

Of the 542 distinct word types in the Bambara dev set, 166 are not contained in the vocabulary (seen during training), 174 of 590 (29.5\%) for the test split. For the French portion it is 243 of 713 (34.1\%) for the dev split and 274 of 756 (36.2\%) for the test split. 
Because of this large proportion of unknown words, we segment the data for both language pairs into subword units (byte pair encodings, BPE) (500 or 1000, separately) using subword-nmt\footnote{\url{https://github.com/rsennrich/subword-nmt}} \citep{sennrich-etal-2016-neural}, and apply BPE dropout to the training sets of both languages \citep{provilkov2019bpedropout}. 
We also experiment with character-level translation for French.

\begin{table*}[t]
    \centering
    \resizebox{\textwidth}{!}{%
    \begin{tabular}{lllllll}
        \toprule
        & & & \multicolumn{2}{c}{\textbf{bam$\rightarrow$fr}} & \multicolumn{2}{c}{\textbf{fr$\rightarrow$bam}} \\
         & \textbf{NMT Model} & \textbf{Configuration} & \textbf{BLEU} & \textbf{ChrF} & \textbf{BLEU} & \textbf{ChrF}\\
        \midrule        
         (1) & Word & 2464 (fr) / 1724 (bam) words & 18.9 & 0.3 & 20.6 & 0.3 \\ 
         (2) & Char & 97 (fr) / 89 (bam) chars & 11.7 & 0.2 & 10.6 & 0.2 \\ 
         
         \midrule
         (2) & BPE & 500 subword merges each & 19.1 & 0.3 & \textbf{21.1} & 0.3 \\ 
         (3) & BPE & 1000 subword merges each & \textbf{19.2} & 0.3 & 20.4 & 0.3 \\ 
         
         \midrule
         (4) & (2) + BPE dropout & dropout=0.1 & 17.8 & 0.3 & 16.6 & 0.2  \\ 
         (5) & (3) + BPE dropout & dropout=0.1 & 16.0 & 0.3 & 17.7 & 0.2 \\ 

         \midrule
         \midrule
         \multicolumn{2}{l}{\textbf{Test scores} for the best models from above} &  & 20.9 & 0.3 & 21.4 & 0.3 \\ 
         \bottomrule
    \end{tabular}%
    }
    \caption{NMT results for French and Bambara translations in corpus BLEU and ChrF on the dev set. The best model (in bold) is evaluated on the test set, results are reported in the last row.}
    \label{tab:results_fr}
\end{table*}

\begin{table*}[t]
    \centering
    \resizebox{\textwidth}{!}{%
    \begin{tabular}{lllllll}
        \toprule
        & & & \multicolumn{2}{c}{\textbf{bam$\rightarrow$en}} & \multicolumn{2}{c}{\textbf{en$\rightarrow$bam}} \\
         & \textbf{NMT Model} & \textbf{Configuration} & \textbf{BLEU} & \textbf{ChrF} & \textbf{BLEU} & \textbf{ChrF}\\
        \midrule        
         (1) & Word & 2364 (en) / 1745 (bam) words & 19.1 & 0.3 & 17.7 & 0.3 \\ 
        (2) & Char & 81 (en) / 86 (bam) chars & 13.2 & 0.2 & 12.5 & 0.2 \\ 
         
        \midrule
         (3) & BPE & 500 subword merges each & 18.1 & 0.3 & \textbf{21.3} & 0.3 \\ 
         (4) & BPE & 1000 subword merges each & \textbf{19.1} & 0.3 & 20.0 & 0.3 \\ 
         
         \midrule
         (5) & (3) + BPE dropout & dropout=0.1 & 17.0 & 0.2 & 19.3 & 0.3 \\ 
         (6) & (4) + BPE dropout & dropout=0.1 & 15.2 & 0.2 & 18.2 & 0.2 \\ 

        \midrule
        \midrule
         \multicolumn{2}{l}{\textbf{Test scores} for the best models from above} &  & 14.8 & 0.3 & 20.9 & 0.3 \\ 
         \bottomrule
    \end{tabular}%
    }
    \caption{NMT results for English and Bambara translations in corpus BLEU and ChrF on the dev set. The best model (in bold) is evaluated on the test set, results are reported in the last row.}
    \label{tab:results_en}
\end{table*}

\section{Results}

\subsection{Automatic Evaluation}
\label{sec:auto-eval}
\paragraph{Segmentation.} We evaluate the models' translations against reference translations on our heldout sets with corpus BLEU \citep{papineni2002bleu} and ChrF \citep{popovic2015chrf} computed with SacreBLEU~\citep{post-2018-call}.\footnote{\url{BLEU+case.mixed+numrefs.1+smooth.exp+tok.intl+version.1.4.9},\\ \url{chrF2+case.mixed+numchars.6+numrefs.1+space.False+version.1.4.9}} Tables~\ref{tab:results_fr} and ~\ref{tab:results_en} show the results for French and English translations respectively. We find that word- and character-level modeling performs sub par compared to subword-level segmentation, which is in line with previous work on low-resource MT. The word-based model cannot resolve out-of-vocabulary words, and the character-level model struggled with word composition. With BPE, smaller subwords seem to perform slightly better than larger ones.
BPE dropout \citep{provilkov2019bpedropout}, which was previously reported to be helpful for low-resource MT ~\citep{richburg-etal-2020-evaluation}, did not increase the quality of the results.
We observe a trend towards higher scores for translations into Bambara than in the reverse direction, but this cross-lingual comparison has to be taken with a grain of salt, since it is influenced by source and target complexity ~\citep{bugliarello-etal-2020-easier}. Ambiguities on the Bambara side, such as the gender of pronouns illustrated in the example in Section~\ref{preprocessing}, might make translation into English and French particularly difficult.

\paragraph{Back-translation.} In addition, we experimented with back-translated Wikipedia data: fine-tuning the original model on a combination of the back-translated and original data, or training it from scratch on a combination of both, as e.g. in ~\citep{przystupa-abdul-mageed-2019-neural}. However, this did not yield improvements over the original BPE model.\footnote{Not included in tables.} We speculate that the mismatch between domains hindered improvement. Indeed, we discovered that when we selected only short sentences from the Wikipedia data, we observed slightly better results, but they still did not outperform the baseline. This highlights the importance of  general domain evaluation sets for future work, so that the effectiveness of leveraging additional out-of-domain data can be measured. 

\paragraph{Multilingual modeling.} Another promising approach for the improvement of extremely low-resourced languages is multilingual modeling ~\citep{johnson-etal-2017-googles}. In our case, we combined the tasks of translating from English and French into Bambara to strengthen the Bambara decoding abilities of the translation model, 
by concatenating the training data and learning joint BPE models. The training data is filtered so that it does not contain sentences from the evaluation sets of the respective other language. 
However, we do not find improvements over the bilingual model.\footnote{Initially, with cross-lingual overlap between training data of one language and evaluation data of the other language we found large improvements, which shows that the model can transfer from English to French sources very well (and vice versa).} We would have expected improvements on translating into Bambara because of larger variation on the source side. However, one reason for not seeing this improvement might be that the sentences are relatively short, and fluency is not as much of an issue as in larger scale studies from previous works.

\subsection{Human evaluation}
 Two native Bambara speakers from Mali, co-authors of this paper, with both college-level French and English reading skills, evaluated a random sample of 21 of the test set translations from Bambara into French and 21 different test set translations from Bambara into English produced by the highest scoring models. Both native speakers received their basic education in Bambara and can read and write the language with fluency.

\paragraph{Evaluation Schema.} For translations that had only a limited correspondence to the source text, the evaluators were given a number of questions specific to the quality of the translations.
For translations that rose to the level of being qualified as conveying most of the sense of the source text we asked for two numerical ratings: First, whether ``most people'' would be able to understand what the meaning of the source sentence from its translation. This was intended to cover translations where the technical accuracy might be low, as a BLEU score might measure, but that substantially conveyed the meaning of the source text. Second, whether the translation was a ``good translation,'' meaning that exact word choices and structure hewed closely to the style and meaning of the source text. We chose to use relatively inexact terminology to describe the ranking criteria as we felt that, as non-professional translators, our evaluators would have difficulty using more technical guidance. 

\paragraph{Quantitative results.} The evaluation schema and the results obtained are presented in Table~\ref{tab:human_eval}. 
\begin{table*}[t]
     \centering
    \resizebox{\textwidth}{!}{%
    \begin{tabular}{llccc}
        \toprule
       & \textbf{Evaluation Criterion} & \textbf{English} & \textbf{French} & \textbf{All } \\
        \midrule
       (1) & Percentage of \emph{words} in the translated sentence that are \emph{related} to the subject of the source. & 43\% & 51\% & 47\% \\
      (2) &   Translated \emph{sentences} containing words \emph{related} to the subject of the source. &  71\% & 52\%	& 62\% \\	
        \midrule
      (3) &  Percentage of \emph{words} that are plausible \emph{direct translations} of words in the source. & 37\% &	40\% &	39\% \\
       (4) & Translated \emph{sentences} containing words that are plausible \emph{direct translations} of words in the source. & 71\% &	52\% & 62\%\\
       \midrule
       (5)& Translated \emph{sentences} that say \emph{something intelligible} whether or not related to the source. & 	71\%	& 66\% & 69\% \\
       (6) & Translated \emph{sentences} that \emph{convey most} of the information in the source.	& 33\% &	38\% &	36\% \\
       (7) & Translated \emph{sentences} where most people would be able to get, at a minimum, \emph{the gist} of the source. & 	33\%	& 38\% & 36\%\\
        \midrule
      (8) &  \emph{Rating} of translation quality for sentences of (7) with \emph{``understandable by most people''} as the criterion. &	4.1/5&	4.8/5 &	4.5/5 \\
       (9) &  \emph{Rating} of translation quality for sentences of (7) with \emph{``good translation''} as the criterion. &	4.4/5 &	4.8/5	& 4.6/5\\
         \bottomrule
    \end{tabular}
    }
    \caption{Evaluation results for translations of the test set from Bambara into English and French. Ratings in the last two lines are on a scale from 1 (poor quality) to 5 (good quality).}
    \label{tab:human_eval}
\end{table*}
The proportion of adequate words appears similar for English and French (rows 1 and 3). However, more English translations are judged as being adequate (rows 2 and 4).  
The overall percentage of translations that might be said to be useful (row 7), in that they convey at least the gist of the source sentence, is low at 36\%, similar to the results obtained by automated methods. 

\paragraph{Examples.}
The following translation excited our admiration because the sentence is relatively complex and the translation is flawless: ``\bam{Farafinna tilebinyanfan tun tilalen b\m{e} Angil\m{e}w ni Tubabuw c\m{e}.}'', which gets translated to ``L'Afrique de l'Ouest a été divisée entre les Anglais et les Français.'' (``West Africa was divided between the English and the French''.)

We also observe that the MT system often translated verb tense correctly, perhaps helped by the fact that verb tenses are extremely simple in Bambara. Some of the sentences that did not qualify as adequate translations nonetheless were instructive and demonstrated specific pattern recognition capabilities.

For another example with the Bambara source
``\bam{I b\m{e} gojogojo wa?}'', the model translates 	``Have you ever eaten you is your wife?''. The word ``gojogojo'' is slang Bambara mainly used by youth playfully employing reduplication, onomatopoeia, and inspiration from a foreign language.  While the term produced a nonsense sentence, the translation seems to carry some of the playfulness in the source sentence. We also notice that it does pick up the subject and uses interrogative word order. In Bambara the word order is the same word order used for a declarative sentence, ``You'' - ``are'' - ``athletic'', the sentence is made interrogative by the  interrogative marker ``\bam{wa}''. 

Looking at the following example, with Bambara source
``\bam{Araba, \m{o}kut\m{o}burukalo tile 8 san 2003}'' and translation	``Après la mort de 25 ans.'', the failure of translating time expressions is surprising because we would have expected the system to have been trained on this pattern - it says, Wednesday, the month of October, the 8th day, the year 2003. The translation (``After the death of 25 years'') is not close. Still, there are markers of time in both the original and translated sentences.

Finally, for Bambara source
``\bam{A nalolen don i n'a f\m{o} suruku.}'' the translation says ``Il met les mains dans ses poches.'' (``He put the hands in his pockets.''), even though the correct translation would say ``He is as crude as a hyena'' (the word for crude does not translate very exactly into English). While the translation seems to have nothing to do with the source, it has the right subject and somehow seems a reasonable guess if you did not know the key words and gives a bit of the spirit of the source sentence.

\section{Conclusion}
Our study constitutes the first attempt of modeling automatic translation for the extremely low-resource language of Bambara. We identified challenges for future work, such as the development of alignment tools for small-scale datasets, and the need for a general domain evaluation set. The current limitation of processing written text as input might furthermore benefit from the integration of spoken resources through speech recognition or speech translation, since Bambara is primarily spoken and the lack of standardization in writing complicates the creation of clean reference sets and consistent evaluation.


\bibliography{anthology,references,aacl-ijcnlp2020}

\begin{thebibliography}{38}
\expandafter\ifx\csname natexlab\endcsname\relax\def\natexlab#1{#1}\fi

\bibitem[{Agi{\'c} and Vuli{\'c}(2019)}]{agic-vulic-2019-jw300}
{\v{Z}}eljko Agi{\'c} and Ivan Vuli{\'c}. 2019.
\newblock \href {https://doi.org/10.18653/v1/P19-1310} {{JW}300: A
  wide-coverage parallel corpus for low-resource languages}.
\newblock In \emph{Proceedings of the 57th Annual Meeting of the Association
  for Computational Linguistics}, pages 3204--3210, Florence, Italy.
  Association for Computational Linguistics.

\bibitem[{Ahia and Ogueji(2020)}]{ahia2020supervised}
Orevaoghene Ahia and Kelechi Ogueji. 2020.
\newblock Towards supervised and unsupervised neural machine translation
  baselines for nigerian pidgin.
\newblock \emph{``AfricaNLP'' Workshop at the 8th International Conference on
  Learning Representations}.

\bibitem[{Akinfaderin(2020)}]{akinfaderin-2020-hausamt}
Adewale Akinfaderin. 2020.
\newblock \href {https://doi.org/10.18653/v1/2020.winlp-1.38} {{H}ausa{MT}
  v1.0: Towards {E}nglish{--}{H}ausa neural machine translation}.
\newblock In \emph{Proceedings of the The Fourth Widening Natural Language
  Processing Workshop}, pages 144--147, Seattle, USA. Association for
  Computational Linguistics.

\bibitem[{Aplonova(2018)}]{aplonova2018development}
Ekaterina Aplonova. 2018.
\newblock Development of a bambara treebank.
\newblock \emph{ARANEA 2018}, page~7.

\bibitem[{Aplonova and Tyers(2017)}]{aplonova2017towards}
Ekaterina Aplonova and Francis Tyers. 2017.
\newblock Towards a dependency-annotated treebank for bambara.
\newblock In \emph{Proceedings of the 16th International Workshop on Treebanks
  and Linguistic Theories}, pages 138--145.

\bibitem[{Arivazhagan et~al.(2019)Arivazhagan, Bapna, Firat, Lepikhin, Johnson,
  Krikun, Chen, Cao, Foster, Cherry, Macherey, Chen, and
  Wu}]{arivazhagan2019massively}
Naveen Arivazhagan, Ankur Bapna, Orhan Firat, Dmitry Lepikhin, Melvin Johnson,
  Maxim Krikun, Mia~Xu Chen, Yuan Cao, George Foster, Colin Cherry, Wolfgang
  Macherey, Zhifeng Chen, and Yonghui Wu. 2019.
\newblock \href {http://arxiv.org/abs/1907.05019} {Massively multilingual
  neural machine translation in the wild: Findings and challenges}.

\bibitem[{Barrault et~al.(2019)Barrault, Bojar, Costa-Juss{\`a}, Federmann,
  Fishel, Graham, Haddow, Huck, Koehn, Malmasi et~al.}]{barrault2019findings}
Lo{\"\i}c Barrault, Ond{\v{r}}ej Bojar, Marta~R Costa-Juss{\`a}, Christian
  Federmann, Mark Fishel, Yvette Graham, Barry Haddow, Matthias Huck, Philipp
  Koehn, Shervin Malmasi, et~al. 2019.
\newblock Findings of the 2019 conference on machine translation (wmt19).
\newblock In \emph{Proceedings of the Fourth Conference on Machine Translation
  (Volume 2: Shared Task Papers, Day 1)}, pages 1--61.

\bibitem[{van Biljon et~al.(2020)van Biljon, Pretorius, and
  Kreutzer}]{biljon2020}
Elan van Biljon, Arnu Pretorius, and Julia Kreutzer. 2020.
\newblock \href {https://arxiv.org/abs/2004.04418} {On optimal transformer
  depth for low-resource language translation}.
\newblock \emph{``AfricaNLP'' Workshop at the 8th International Conference on
  Learning Representations}.

\bibitem[{Bugliarello et~al.(2020)Bugliarello, Mielke, Anastasopoulos,
  Cotterell, and Okazaki}]{bugliarello-etal-2020-easier}
Emanuele Bugliarello, Sabrina~J. Mielke, Antonios Anastasopoulos, Ryan
  Cotterell, and Naoaki Okazaki. 2020.
\newblock \href {https://doi.org/10.18653/v1/2020.acl-main.149} {It{'}s easier
  to translate out of {E}nglish than into it: {M}easuring neural translation
  difficulty by cross-mutual information}.
\newblock In \emph{Proceedings of the 58th Annual Meeting of the Association
  for Computational Linguistics}, pages 1640--1649, Online. Association for
  Computational Linguistics.

\bibitem[{Culy(1985)}]{culy1985complexity}
Christopher Culy. 1985.
\newblock The complexity of the vocabulary of bambara.
\newblock \emph{Linguistics and philosophy}, 8(3):345--351.

\bibitem[{Dossou and Emezue(2020)}]{dossou2020ffr}
Bonaventure F.~P. Dossou and Chris~C. Emezue. 2020.
\newblock \href {http://arxiv.org/abs/2003.12111} {Ffr v1.0: Fon-french neural
  machine translation}.

\bibitem[{{$\forall$} et~al.(2020){$\forall$}, Orife, Kreutzer, Sibanda,
  Whitenack, Siminyu, Martinus, Ali, Abbott, Marivate, Kabongo, Meressa,
  Murhabazi, Ahia, van Biljon, Ramkilowan, Akinfaderin, Öktem, Akin, Kioko,
  Degila, Kamper, Dossou, Emezue, Ogueji, and Bashir}]{masakhane}
{ } {$\forall$}, Iroro Orife, Julia Kreutzer, Blessing Sibanda, Daniel
  Whitenack, Kathleen Siminyu, Laura Martinus, Jamiil~Toure Ali, Jade Abbott,
  Vukosi Marivate, Salomon Kabongo, Musie Meressa, Espoir Murhabazi,
  Orevaoghene Ahia, Elan van Biljon, Arshath Ramkilowan, Adewale Akinfaderin,
  Alp Öktem, Wole Akin, Ghollah Kioko, Kevin Degila, Herman Kamper,
  Bonaventure Dossou, Chris Emezue, Kelechi Ogueji, and Abdallah Bashir. 2020.
\newblock \href {https://arxiv.org/pdf/2003.11529} {Masakhane -- machine
  translation for africa}.

\bibitem[{Goyal et~al.(2020)Goyal, Kumar, and
  Sharma}]{goyal-etal-2020-efficient}
Vikrant Goyal, Sourav Kumar, and Dipti~Misra Sharma. 2020.
\newblock \href {https://doi.org/10.18653/v1/2020.acl-srw.22} {Efficient neural
  machine translation for low-resource languages via exploiting related
  languages}.
\newblock In \emph{Proceedings of the 58th Annual Meeting of the Association
  for Computational Linguistics: Student Research Workshop}, pages 162--168,
  Online. Association for Computational Linguistics.

\bibitem[{Groenewald and du~Plooy(2010)}]{autshumato}
J.~Hendrik Groenewald and Liza du~Plooy. 2010.
\newblock Processing parallel text corpora for three south african language
  pairs in the autshumato project.
\newblock In \emph{Proceedings of the Second Workshop on African Language
  Technology}, Valletta, Malta.

\bibitem[{Heine and Nurse(2000)}]{heine2000african}
Bernd Heine and Derek Nurse. 2000.
\newblock \emph{African languages: An introduction}.
\newblock Cambridge University Press.

\bibitem[{Johnson(2018)}]{johnson_2018}
Melvin Johnson. 2018.
\newblock \href
  {https://ai.googleblog.com/2018/12/providing-gender-specific-translations.html}
  {Providing gender-specific translations in google translate}.

\bibitem[{Johnson et~al.(2017)Johnson, Schuster, Le, Krikun, Wu, Chen, Thorat,
  Vi{\'e}gas, Wattenberg, Corrado, Hughes, and
  Dean}]{johnson-etal-2017-googles}
Melvin Johnson, Mike Schuster, Quoc~V. Le, Maxim Krikun, Yonghui Wu, Zhifeng
  Chen, Nikhil Thorat, Fernanda Vi{\'e}gas, Martin Wattenberg, Greg Corrado,
  Macduff Hughes, and Jeffrey Dean. 2017.
\newblock \href {https://doi.org/10.1162/tacl_a_00065} {{G}oogle{'}s
  multilingual neural machine translation system: Enabling zero-shot
  translation}.
\newblock \emph{Transactions of the Association for Computational Linguistics},
  5:339--351.

\bibitem[{Kingma and Ba(2014)}]{kingma2014adam}
Diederik~P Kingma and Jimmy Ba. 2014.
\newblock Adam: A method for stochastic optimization.
\newblock \emph{arXiv preprint arXiv:1412.6980}.

\bibitem[{Kreutzer et~al.(2019)Kreutzer, Bastings, and
  Riezler}]{kreutzer-etal-2019-joey}
Julia Kreutzer, Jasmijn Bastings, and Stefan Riezler. 2019.
\newblock \href {https://doi.org/10.18653/v1/D19-3019} {Joey {NMT}: A
  minimalist {NMT} toolkit for novices}.
\newblock In \emph{Proceedings of the 2019 Conference on Empirical Methods in
  Natural Language Processing and the 9th International Joint Conference on
  Natural Language Processing (EMNLP-IJCNLP): System Demonstrations}, pages
  109--114, Hong Kong, China. Association for Computational Linguistics.

\bibitem[{Leventhal et~al.(2020)Leventhal, Tapo, Luger, Zampieri, and
  Homan}]{leventhal2020assessing}
Michael Leventhal, Allahsera Tapo, Sarah Luger, Marcos Zampieri, and
  Christopher~M. Homan. 2020.
\newblock \href {http://arxiv.org/abs/2004.00068} {Assessing human translations
  from french to bambara for machine learning: a pilot study}.

\bibitem[{Lewis et~al.(2014)Lewis, Simons, and Fennig}]{lewis2014ethnologue}
M~Paul Lewis, Gary~F Simons, and Charles~D Fennig. 2014.
\newblock \emph{Ethnologue: Languages of Africa and Europe}.
\newblock SIL international.

\bibitem[{Luger et~al.(2020)Luger, Tapo, Homan, Zampieri, and
  Leventhal}]{luger2020}
Sarah Luger, Allashera~Auguste Tapo, Christopher~M. Homan, Marcos Zampieri, and
  Michael Leventhal. 2020.
\newblock Towards a crowdsourcing platform for low resource languages -- a
  semi-supervised approach.
\newblock In \emph{Proceedings of the Eighth Conference on Human Computations}.
  AAAI.

\bibitem[{Martinus and Abbott(2019)}]{DBLP:journals/corr/abs-1906-05685}
Laura Martinus and Jade~Z. Abbott. 2019.
\newblock \href {http://arxiv.org/abs/1906.05685} {A focus on neural machine
  translation for african languages}.
\newblock \emph{CoRR}, abs/1906.05685.

\bibitem[{Orife(2020)}]{orife2020neural}
Iroro Orife. 2020.
\newblock Towards neural machine translation for edoid languages.
\newblock \emph{``AfricaNLP'' Workshop at the 8th International Conference on
  Learning Representations}.

\bibitem[{Papineni et~al.(2002)Papineni, Roukos, Ward, and
  Zhu}]{papineni2002bleu}
Kishore Papineni, Salim Roukos, Todd Ward, and Wei-Jing Zhu. 2002.
\newblock Bleu: a method for automatic evaluation of machine translation.
\newblock In \emph{Proceedings of the 40th annual meeting on association for
  computational linguistics}, pages 311--318. Association for Computational
  Linguistics.

\bibitem[{Paszke et~al.(2019)Paszke, Gross, Massa, Lerer, Bradbury, Chanan,
  Killeen, Lin, Gimelshein, Antiga, Desmaison, Kopf, Yang, DeVito, Raison,
  Tejani, Chilamkurthy, Steiner, Fang, Bai, and Chintala}]{NIPS2019_9015}
Adam Paszke, Sam Gross, Francisco Massa, Adam Lerer, James Bradbury, Gregory
  Chanan, Trevor Killeen, Zeming Lin, Natalia Gimelshein, Luca Antiga, Alban
  Desmaison, Andreas Kopf, Edward Yang, Zachary DeVito, Martin Raison, Alykhan
  Tejani, Sasank Chilamkurthy, Benoit Steiner, Lu~Fang, Junjie Bai, and Soumith
  Chintala. 2019.
\newblock \href
  {http://papers.nips.cc/paper/9015-pytorch-an-imperative-style-high-performance-deep-learning-library.pdf}
  {Pytorch: An imperative style, high-performance deep learning library}.
\newblock In H.~Wallach, H.~Larochelle, A.~Beygelzimer, F.~d\textquotesingle
  Alch\'{e}-Buc, E.~Fox, and R.~Garnett, editors, \emph{Advances in Neural
  Information Processing Systems 32}, pages 8026--8037. Curran Associates, Inc.

\bibitem[{Popovi{\'c}(2015)}]{popovic2015chrf}
Maja Popovi{\'c}. 2015.
\newblock chrf: character n-gram f-score for automatic mt evaluation.
\newblock In \emph{Proceedings of the Tenth Workshop on Statistical Machine
  Translation}, pages 392--395.

\bibitem[{Post(2018)}]{post-2018-call}
Matt Post. 2018.
\newblock \href {https://doi.org/10.18653/v1/W18-6319} {A call for clarity in
  reporting {BLEU} scores}.
\newblock In \emph{Proceedings of the Third Conference on Machine Translation:
  Research Papers}, pages 186--191, Belgium, Brussels. Association for
  Computational Linguistics.

\bibitem[{Provilkov et~al.(2019)Provilkov, Emelianenko, and
  Voita}]{provilkov2019bpedropout}
Ivan Provilkov, Dmitrii Emelianenko, and Elena Voita. 2019.
\newblock \href {http://arxiv.org/abs/1910.13267} {Bpe-dropout: Simple and
  effective subword regularization}.

\bibitem[{Przystupa and
  Abdul-Mageed(2019)}]{przystupa-abdul-mageed-2019-neural}
Michael Przystupa and Muhammad Abdul-Mageed. 2019.
\newblock \href {https://doi.org/10.18653/v1/W19-5431} {Neural machine
  translation of low-resource and similar languages with backtranslation}.
\newblock In \emph{Proceedings of the Fourth Conference on Machine Translation
  (Volume 3: Shared Task Papers, Day 2)}, pages 224--235, Florence, Italy.
  Association for Computational Linguistics.

\bibitem[{Richburg et~al.(2020)Richburg, Eskander, Muresan, and
  Carpuat}]{richburg-etal-2020-evaluation}
Aquia Richburg, Ramy Eskander, Smaranda Muresan, and Marine Carpuat. 2020.
\newblock \href {https://doi.org/10.18653/v1/2020.winlp-1.40} {An evaluation of
  subword segmentation strategies for neural machine translation of
  morphologically rich languages}.
\newblock In \emph{Proceedings of the The Fourth Widening Natural Language
  Processing Workshop}, pages 151--155, Seattle, USA. Association for
  Computational Linguistics.

\bibitem[{Sennrich et~al.(2016)Sennrich, Haddow, and
  Birch}]{sennrich-etal-2016-neural}
Rico Sennrich, Barry Haddow, and Alexandra Birch. 2016.
\newblock \href {https://doi.org/10.18653/v1/P16-1162} {Neural machine
  translation of rare words with subword units}.
\newblock In \emph{Proceedings of the 54th Annual Meeting of the Association
  for Computational Linguistics (Volume 1: Long Papers)}, pages 1715--1725,
  Berlin, Germany. Association for Computational Linguistics.

\bibitem[{Szegedy et~al.(2016)Szegedy, Vanhoucke, Ioffe, Shlens, and
  Wojna}]{szegedy2016rethinking}
Christian Szegedy, Vincent Vanhoucke, Sergey Ioffe, Jon Shlens, and Zbigniew
  Wojna. 2016.
\newblock Rethinking the inception architecture for computer vision.
\newblock In \emph{Proceedings of the IEEE conference on computer vision and
  pattern recognition}, pages 2818--2826.

\bibitem[{Vaswani et~al.(2017)Vaswani, Shazeer, Parmar, Uszkoreit, Jones,
  Gomez, Kaiser, and Polosukhin}]{vaswani2017attention}
Ashish Vaswani, Noam Shazeer, Niki Parmar, Jakob Uszkoreit, Llion Jones,
  Aidan~N Gomez, {\L}ukasz Kaiser, and Illia Polosukhin. 2017.
\newblock Attention is all you need.
\newblock In \emph{Advances in Neural Information Processing Systems
  {(NeurIPS)}}, Long Beach, {CA, USA}.

\bibitem[{Vydrin(2018)}]{vydrin2018corpus}
Valentin Vydrin. 2018.
\newblock \href {https://doi.org/10.31862/2500-2953-2018-4-34-49} {Where corpus
  methods hit their limits: the case of separable adjectives in bambara}.
\newblock \emph{{Rhema}}, (4).

\bibitem[{Vydrin et~al.(2011)Vydrin, Maslinsky, M{\'e}ric, and
  Rovenchak}]{vydrin2011corpus}
Valentin Vydrin, Kirill Maslinsky, Jean-Jacques M{\'e}ric, and A~Rovenchak.
  2011.
\newblock Corpus bambara de r{\'e}f{\'e}rence.

\bibitem[{Wu et~al.(2019)Wu, DeMattos, So, Chen, and
  {\c{C}}{\"o}ltekin}]{wu2019language}
Nianheng Wu, Eric DeMattos, Kwok~Him So, Pin-zhen Chen, and
  {\c{C}}a{\u{g}}r{\i} {\c{C}}{\"o}ltekin. 2019.
\newblock Language discrimination and transfer learning for similar languages:
  experiments with feature combinations and adaptation.
\newblock In \emph{Proceedings of the Sixth Workshop on NLP for Similar
  Languages, Varieties and Dialects}, pages 54--63.

\bibitem[{Öktem et~al.(2020)Öktem, Plitt, and Tang}]{ktem2020tigrinya}
Alp Öktem, Mirko Plitt, and Grace Tang. 2020.
\newblock Tigrinya neural machine translation with transfer learning for
  humanitarian response.
\newblock \emph{``AfricaNLP'' Workshop at the 8th International Conference on
  Learning Representations}.

\end{thebibliography}
\bibliographystyle{acl_natbib}

\appendix


\section{Appendices}
\label{sec:appendix}

%
\begin{landscape}
\begin{table}[!h]
\scalebox{0.70}{
\begin{tabular}{|c|c|c|c|c|}
\hline
\textbf{URL} &
  \textbf{Description} &
  \textbf{Rate} &
  \textbf{Pros} &
  \textbf{Cons} \\ \hline
\url{http://www.alanwood.net/unicode/n\%27ko.html} &
  \begin{tabular}[c]{@{}l@{}}Test for Unicode support \\ in web browsers\end{tabular} &
  2 &
  \begin{tabular}[c]{@{}l@{}}character(n'ko, decimal),\\  decimal, character(n'ko, \\ hex), hex, name.\end{tabular} &
  \begin{tabular}[c]{@{}l@{}}no audio present, no \\ usable data.\end{tabular} \\ \hline
\url{http://www.fakoli.net/index.html} &
  \begin{tabular}[c]{@{}l@{}}N'ko Institute of America,\\  dedicated to the \\ exploration of n’ko\end{tabular} &
  2 &
  \begin{tabular}[c]{@{}l@{}}accessible to English \\ speakers.\end{tabular} &
  \begin{tabular}[c]{@{}l@{}}confusing, writes n’ko\\ using latin alphabets, no \\ audio, and no usable data.\end{tabular} \\ \hline
\begin{tabular}[c]{@{}l@{}}\url{http://www.personal.psu.edu/ejp10/symbolcodes/bylanguage/nko.html}\end{tabular} &
  \begin{tabular}[c]{@{}l@{}}N'ko computing information\\  (penn state)\end{tabular} &
  2 &
  \begin{tabular}[c]{@{}l@{}}facilitate installation of \\ n’ko fonts and gives \\ general information of it\\  as a font\end{tabular} &
  \begin{tabular}[c]{@{}l@{}}no usable data.\end{tabular} \\ \hline
\begin{tabular}[c]{@{}l@{}}\url{https://ipfs.io/ipfs/QmXoypizjW3WknFiJnKLwHCnL72vedxjQkDDP1mXWo6uco/wiki/N\%27Ko_alphabet.html}\end{tabular} &
  N'ko alphabet &
  2 &
  \begin{tabular}[c]{@{}l@{}}provides information about the \\ alphabet from all angle possible\end{tabular} &
  \begin{tabular}[c]{@{}l@{}}no audio present, no usable \\ data.\end{tabular} \\ \hline
\url{http://www.bmanuel.org/clr/clr2_mp.html} &
  \begin{tabular}[c]{@{}l@{}}Multilingual and parallel\\ corpora\end{tabular} &
  1 &
  n/a &
  \begin{tabular}[c]{@{}l@{}}Unaccessible when we \\ recently visited. no usable \\ data.\end{tabular} \\ \hline
\begin{tabular}[c]{@{}l@{}}\url{https://podcasts.apple.com/us/podcast/learn-bambara-with-jeff-frazee/id433868610}\end{tabular} &
  podcast &
  2 &
  short speeches. &
  \begin{tabular}[c]{@{}l@{}}not accurate, no usable \\ data.\end{tabular} \\ \hline
\url{http://www.rfi.fr/emission/kan-jum-be-yen-mandenkan} &
  rss/podcast &
  2 &
  audio available &
  \begin{tabular}[c]{@{}l@{}}not transcribed, no usable \\ data.\end{tabular} \\ \hline
\url{https://www.voabambara.com/podcasts} &
  rss/podcast &
  2 &
  audio/video available &
  \begin{tabular}[c]{@{}l@{}}not transcribed, no usable\\ data.\end{tabular} \\ \hline
\url{https://www.voaafrique.com/z/3551} &
  podcast &
  2 &
  audio/video available &
  \begin{tabular}[c]{@{}l@{}}not transcribed, no usable \\ data.\end{tabular} \\ \hline
\url{https://www.voabambara.com/z/5098} &
  podcast &
  2 &
  audio/video available &
  \begin{tabular}[c]{@{}l@{}}not transcribed, no usable \\ data.\end{tabular} \\ \hline
\url{http://ma.rfi.fr/} &
  streaming station &
  2 &
  available &
  \begin{tabular}[c]{@{}l@{}}not transcribed, no usable \\ data.\end{tabular} \\ \hline
\url{http://dictionary.ankataa.com} &
  dictionary &
  2 &
  available; easily understandable &
  \begin{tabular}[c]{@{}l@{}}no phonetics, no usable \\ data.\end{tabular} \\ \hline
\url{http://www.mali-pense.net/bm/lexicon/index.htm} &
  dictionary &
  2 &
  french/bambara – bambara/french &
  \begin{tabular}[c]{@{}l@{}}no supporting audio, no \\ usable data.\end{tabular} \\ \hline
\url{http://cormand.huma-num.fr/} &
  \begin{tabular}[c]{@{}l@{}}project of bambara de\\  reference\end{tabular} &
  3 &
  Dictionary like &
  Not publicly available. \\ \hline
\end{tabular}}

\caption[Dataset discovery information.]{Datasets for Bambara including: a URL, linking to where we found the source; a description, describing what it is about (as we understood it); a rating from 1 to 5, one is ``related to our topic with no usable data'', and five is ``related to our topic with usable data''; pros, describing its pluses; and cons, describing its minuses.}
\label{tab:data_sources}
\end{table}
\end{landscape}


\end{document}